%% file: main.tex
\documentclass{article} % For LaTeX2e
\usepackage{iclr2021_conference,times}

\input{math_commands.tex}

\usepackage{hyperref}
\usepackage{url}
\usepackage{graphicx}

\usepackage{tikz}
\usepackage{blindtext}
\usepackage{hyperref}
\usepackage{url}
\usepackage{float}
\usepackage{multirow}
\usepackage{booktabs}
\usepackage{caption}
\usepackage{comment}
\usepackage[titlenumbered,ruled]{algorithm2e}
\usepackage{algorithmic}

\usepackage{array}
\usepackage{wrapfig}
\usepackage{fdsymbol}

\usepackage{wrapfig}
\usepackage{subcaption}

\usepackage{microtype}
\usepackage{graphicx}
\usepackage{booktabs} % for professional tables

\usepackage{hhline}
\usepackage{textcomp}

\newcommand{\tabincell}[2]{\begin{tabular}{@{}#1@{}}#2\end{tabular}} %

\definecolor{Note_color}{rgb}{0.0, 0.0, 1.0}

\title{CPT: Efficient Deep Neural Network Training \centering  via Cyclic Precision}

\author{Yonggan Fu, Han Guo, Xin Yang, Yining Ding \& Yingyan Lin
\\
Department of Electrical and Computer Engineering\\
Rice University\\
\texttt{\{yf22, hg31, xy33, yd31, yingyan.lin\}@rice.edu} \\
\And
Meng Li \& Vikas Chandra \\
Facebook Inc \\
\texttt{\{meng.li, vchandra\}@fb.com} \\
}

\iclrfinalcopy % Uncomment for camera-ready version, but NOT for submission.
\begin{document}

\maketitle

\input{Sections/0-Abstract}
\input{Sections/1-Introduction}

\input{Sections/2-Related_work}
\input{Sections/3-Method}
\input{Sections/4-Experiment}
\input{Sections/5-Conclusion}

\bibliography{iclr2021_conference}
\bibliographystyle{iclr2021_conference}

\input{Sections/6-Appendix}

\end{document}

%% file: math_commands.tex
\usepackage{amsmath,amsfonts,bm}

\def\eqref#1{equation~\ref{#1}}
\def\1{\bm{1}}

\DeclareMathAlphabet{\mathsfit}{\encodingdefault}{\sfdefault}{m}{sl}
\SetMathAlphabet{\mathsfit}{bold}{\encodingdefault}{\sfdefault}{bx}{n}

%% file: Sections/0-Abstract.tex
\begin{abstract}

% Abstract for registration
Low-precision deep neural network (DNN) training has gained tremendous attention as reducing precision is one of the most effective knobs for boosting DNNs' training time/energy efficiency. In this paper, we attempt to explore low-precision training from a new perspective as inspired by recent findings in understanding DNN training: we conjecture that DNNs' precision might have a similar effect as the learning rate during DNN training, and advocate dynamic precision along the training trajectory for further boosting the time/energy efficiency of DNN training. Specifically, we propose Cyclic Precision Training (CPT) to cyclically vary the precision between two boundary values which can be identified using a simple precision range test within the first few training epochs. Extensive simulations and ablation studies on five datasets and eleven models demonstrate that CPT's effectiveness is consistent across various models/tasks (including classification and language modeling). Furthermore, through experiments and visualization we show that CPT helps to (1) converge to a wider minima with a lower generalization error and (2) reduce training variance which we believe opens up a new design knob for simultaneously improving the optimization and efficiency of DNN training. Our codes are available at: \href{https://github.com/RICE-EIC/CPT}{https://github.com/RICE-EIC/CPT}.

\end{abstract}

%% file: Sections/1-Introduction.tex
\vspace{-1.2em}
\section{Introduction}
\vspace{-0.5em}

The record-breaking performance of modern deep neural networks (DNNs) comes at a prohibitive training cost due to the required massive training data and parameters, limiting the development of the highly demanded DNN-powered intelligent solutions for numerous applications \citep{AdaDeep,wu2018deep}. As an illustration, training ResNet-50 involves $10^{18}$ FLOPs (floating-point operations) and can take 14 days on one state-of-the-art (SOTA) GPU \citep{you2020drawing}. Meanwhile, the large DNN training costs have raised increasing financial and environmental concerns. For example, it is estimated that training one DNN can cost more than \$10K US dollars and emit carbon as high as a car's lifetime emissions. In parallel, recent DNN advances have fueled a tremendous need for intelligent edge devices, many of which require on-device in-situ learning to ensure the accuracy under dynamic real-world environments, where there is a mismatch between the devices' limited resources and the prohibitive training costs \citep{wang2019e2,li2020halo,you2020shiftaddnet}.  

To address the aforementioned challenges, extensive research efforts have been devoted to developing efficient DNN training techniques. Among them, low-precision training has gained significant attention as it can largely boost the training time/energy efficiency \citep{jacob2018quantization,wang2018training,sun2019hybrid}. For instance, GPUs can now perform mixed-precision DNN training with 16-bit IEEE Half-Precision floating-point formats \citep{mixed-precision}. Despite their promise, existing low-precision works have not yet fully explored the opportunity of leveraging recent findings in understanding DNN training.
In particular, existing works mostly fix the model precision during the whole training process, i.e., adopt a \textbf{static} quantization strategy, while recent works in DNN training optimization suggest dynamic hyper-parameters along DNNs' training trajectory. For example, ~\citep{li2019towards} shows that a large initial learning rate helps the model to memorize easier-to-fit and more generalizable patterns, which aligns with the common practice to start from a large learning rate for exploration and anneal to a small one for final convergence; and~\citep{smith2017cyclical, loshchilov2016sgdr} improve DNNs' classification accuracy by adopting cyclical learning rates.

In this work, we advocate dynamic precision training, and make the following contributions: 
 \vspace{-0.4em}
\begin{itemize}
\item We show that DNNs' precision seems to have a similar effect as the learning rate during DNN training, i.e., low precision with large quantization noise helps DNN training exploration while high precision with more accurate updates aids model convergence, and dynamic precision schedules help DNNs converge to a better minima.
% smooth the loss surface of DNN training. 
This finding opens up a design knob for simultaneously improving the optimization and efficiency of DNN training. 

 \vspace{-0.1em}
\item We propose Cyclic Precision Training (CPT) which adopts a cyclic precision schedule along DNNs' training trajectory for pushing forward the achievable trade-offs between DNNs' accuracy and training efficiency. Furthermore, we show that the cyclic precision bounds can be automatically identified at the very early stage of training using a simple precision range test, which has a negligible computational overhead.
 \vspace{-0.1em}

\item Extensive experiments on \textbf{\textcolor{black}{five} datasets} and \textbf{\textcolor{black}{eleven} models} across a wide spectrum of applications (including classification
% , object detection, 
% \vspace{-0.5em}
and language modeling) validate the consistent effectiveness of the proposed CPT technique in boosting the training efficiency while leading to a comparable or even better accuracy. Furthermore, we provide loss surface visualization for better understanding CPT's effectiveness and discuss its connection with recent findings in understanding DNNs' training optimization.

% \item Extensive experiments on \textbf{six datasets} and \textbf{eleven models} across a wide spectrum of applications (including classification, object detection, and natural language processing) validate the consistent effectiveness of the proposed CPT technique, e.g., \textcolor{red}{ xx\% higher accuracy with xx\% reduced training latency/energy/Bitops}. Furthermore, we provide loss surface visualization for better understanding CPT's effectiveness and discuss its connection with recent findings about DNN training optimization.

% Cyclic Precision Training consistently achieve higher accuracy while reducing training cost. In particular, \textcolor{blue}{(impressive results.)}     
 
\end{itemize}

%% file: Sections/2-Related_work.tex
\vspace{-1.2em}
\section{Related works}
\vspace{-0.5em}
\textbf{Quantized DNNs.}
DNN quantization~\citep{courbariaux2015binaryconnect, courbariaux2016binarized, rastegari2016xnor,zhu2016trained, li2016ternary,jacob2018quantization, mishra2017apprentice, mishra2017wrpn, park2017weighted, zhou2016dorefa} 
% to 1-bit~\citep{courbariaux2015binaryconnect, courbariaux2016binarized, rastegari2016xnor} 
% or higher fixed-point precision~\citep{zhu2016trained, li2016ternary,jacob2018quantization, mishra2017apprentice, mishra2017wrpn, park2017weighted, zhou2016dorefa} 
has been well explored based on the target accuracy-efficiency trade-offs. For example, ~\citep{jacob2018quantization} proposes quantization-aware training
to preserve the post quantization accuracy; ~\citep{jung2019learning, bhalgat2020lsq+, esser2019learned, park2020profit} strive to improve low-precision DNNs' accuracy using learnable quantizers. 
Mixed-precision DNN quantization~\citep{wang2019haq, xu2018dnq, elthakeb2020releq, zhou2017adaptive} assigns different bitwidths for different layers/filters.
While these works all adopt a \textbf{static} quantization strategy, i.e., the assigned precision is fixed post quantization, CPT adopts a \textbf{dynamic} precision schedule during the training process.

\textbf{Low-precision DNN training.}
Pioneering works~\citep{wang2018training, banner2018scalable, micikevicius2017mixed, gupta2015deep, sun2019hybrid} have shown that DNNs can be trained with reduced precision.
For distributed learning, 
~\citep{seide20141, de2017understanding, wen2017terngrad, bernstein2018signsgd} quantize the gradients to reduce the communication costs, where the training computations still
adopt full precision; For centralized/on-device learning,
the weights, activations, gradients, and errors involved in both the forward and backward computations all adopt reduced precision.
% Specifically,~\citep{wang2018training, sun2019hybrid, micikevicius2017mixed, mellempudi2019mixed} adopt reduced-precision floating-point formats, while~\citep{gupta2015deep, banner2018scalable, zhou2016dorefa, yang2020training, wu2018training} use fixed-point representations.
Our CPT can be applied on top of these low-precision training techniques, all of which adopt a \textbf{static} precision during the \textbf{whole} training trajectory, to further boost the training efficiency.

\textbf{Dynamic-precision DNNs.}
There exist some dynamic precision works which aim to derive a quantized DNN for inference after the full-precision training. 
Specifically,~\citep{zhuang2018towards} first trains a full-precision model to reach convergence and then gradually decreases the model precision to the target one for achieving better inference accuracy; ~\citep{khoram2018adaptive} also starts from a full-precision model and then gradually learns the precision of each layer to derive a mixed-precision counterpart; ~\citep{yang2020fracbits} learns a fractional precision of each layer/filter based on the linear interpolation of two consecutive bitwidths which doubles the computation and requires an extra fine-tuning step; and \citep{shen2020fractional} proposes to adapt the precision of each layer during inference in an input-dependent manner to balance computational cost and accuracy.

%% file: Sections/3-Method.tex
\vspace{-0.5em}
\section{The Proposed CPT Technique}
\vspace{-0.5em}
In this section, we first introduce the hypothesis that motivates us to develop CPT using visualization examples in Sec.~\ref{sec:example}, and then present the CPT concept in Sec.~\ref{sec:cyclic_precsion} followed by the Precision Range Test (PRT) method in Sec.~\ref{sec:range_test}, where PRT aims to automate the precision schedule for CPT.

% In this section, we start from the analysis about the role of precision in DNN training with a motivating example in Sec.~\ref{sec:example}.Then we propose Cyclic Precision Training (CPT) in Sec.~\ref{sec:cyclic_precsion} with the Precision Range Test (PRT) technique in Sec.~\ref{sec:range_test} to fully automate the precision schedule.

\vspace{-1em}
\subsection{CPT: Motivation}
\label{sec:example}
\vspace{-0.5em}

% \textbf{Motivation hypothesis.} Recent findings in DNN training have motivated us to rethink the role of DNN precision. Specifically, it has been discussed that (1) DNNs \textbf{learn to fit different patterns at different training stages}, e.g., \citep{pmlr-v97-rahaman19a, xu2019frequency} reveals that DNN training first learns lower-frequency components and then high-frequency features, with the former being more robust to perturbations and noises; and (2) \textbf{dynamic learning rate schedules help} to improve the optimization in DNN training, e.g., ~\citep{li2019towards} points out that a large initial learning rate helps the model to memorize easier-to-fit and more generalizable patterns while ~\citep{smith2017cyclical, loshchilov2016sgdr} shows that cyclical learning rate schedules improve DNNs' classification accuracy. These works inspire us to \textbf{hypothesize} that precision has a similar effect as the learning rate and dynamic precision might help DNNs to reach a better optimum in the optimization landscape.

\textbf{Hypothesis 1: DNN's precision has a similar effect as the learning rate.} Existing works~\citep{grandvalet1997noise, neelakantan2015adding} show that noise can help DNN training theoretically or empirically, motivating us to rethink the role of quantization in DNN training. We conjecture that low precision with large quantization noise helps DNN training exploration with an effect similar to a high learning rate, while high precision with more accurate updates aids model convergence, similar to a low learning rate.

\textbf{Validating Hypothesis 1.} \underline{Settings:} To empirically justify our hypothesis, we train ResNet-38/74 on the CIFAR-100 dataset for 160 epochs following the basic training setting as in Sec.~\ref{sec:exp_setup}. In particular, we divide the training of 160 epochs into three stages: [0-th, 80-th], [80-th,120-th], and [120-th, 160-th]: for the first training stage of [0-th, 80-th], we adopt different learning rates and precisions for the weights and activations, while using full precision for the remaining two stages with a learning rate of 0.01 for the [80-th,120-th] epochs and 0.001 for the [120-th, 160-th] epochs in all the experiments in order to explore the relationship between the learning rate and precision in the first training stage.

\begin{table}[t]
\centering
\vspace{-1em}
\caption{The test accuracy of ResNet-38/74 trained on CIFAR-100 with different learning rate and precision combinations in the first stage. Note that the last two stages of all the experiments are trained with full precision and a learning rate of 0.01 and 0.001, respectively.}
\vspace{-0.5em}
\resizebox{0.8\textwidth}{!}{
\begin{tabular}{ccccccccc}
\hline
 & \multicolumn{4}{c}{\textbf{ResNet-38}} & \multicolumn{4}{c}{\textbf{ResNet-74}} \\ \hline
First-stage LR & 0.1 & 0.06 & 0.03 & 0.01  & 0.1 & 0.06 & 0.03 & 0.01 \\ \hline
4-bit Acc (\%) & 69.45 & 68.63 & \textbf{67.69} & 65.90  & 70.96 & 69.54 & 68.26 & \textbf{67.19} \\
6-bit Acc (\%) & 70.22 & 68.87 & 67.15 & \textbf{66.10}  & 71.62 & 70.28 & \textbf{68.84} & 66.16 \\
8-bit Acc (\%) & 69.96 & 68.66 & 66.75 & 64.99 & 71.60 & \textbf{70.67} & 68.45 & 65.85 \\
FP Acc (\%) & \textbf{70.45} & \textbf{69.53} & 67.47 & 64.50 & \textbf{71.66} & 70.00 & 68.69 & 65.62 \\ \hline
\end{tabular}
}
\label{tab:lr-prec}
% \vspace{-2em}
\end{table}

\underline{Results:} As shown in Tab.~\ref{tab:lr-prec}, we can observe that as the learning rate is sufficiently reduced for the first training stage, adopting a lower precision for this stage will lead to a higher accuracy than training with full precision. In particular, with the standard initial learning rate of 0.1, full precision training achieves a 1.00\%/0.70\% higher accuracy than the 4-bit one on ResNet-38/74, respectively; whereas as the initial learning rate decreases, this accuracy gap gradually narrows and then reverses, e.g., when the initial learning rate becomes 1e-2, training with [0-th, 80-th] of 4-bit achieves a 1.40\%/1.57\% higher accuracy than the full precision ones. 

\underline{Insights:} This set of experiments show that (1) when the initial learning rate is low, training with lower initial precisions consistently leads to a better accuracy than training with full precision, indicating that lowering the precision introduces a similar effect  of favoring exploration as that of a high learning rate; and (2) although a low precision can alleviate the accuracy drop caused by a low learning rate, a high learning rate is in general necessary to maximize the accuracy.

\begin{wrapfigure}{r}{0.45\textwidth}
% \begin{figure*}[!t]
    \centering
    \vspace{-2em}
    \includegraphics[width=0.45\textwidth]{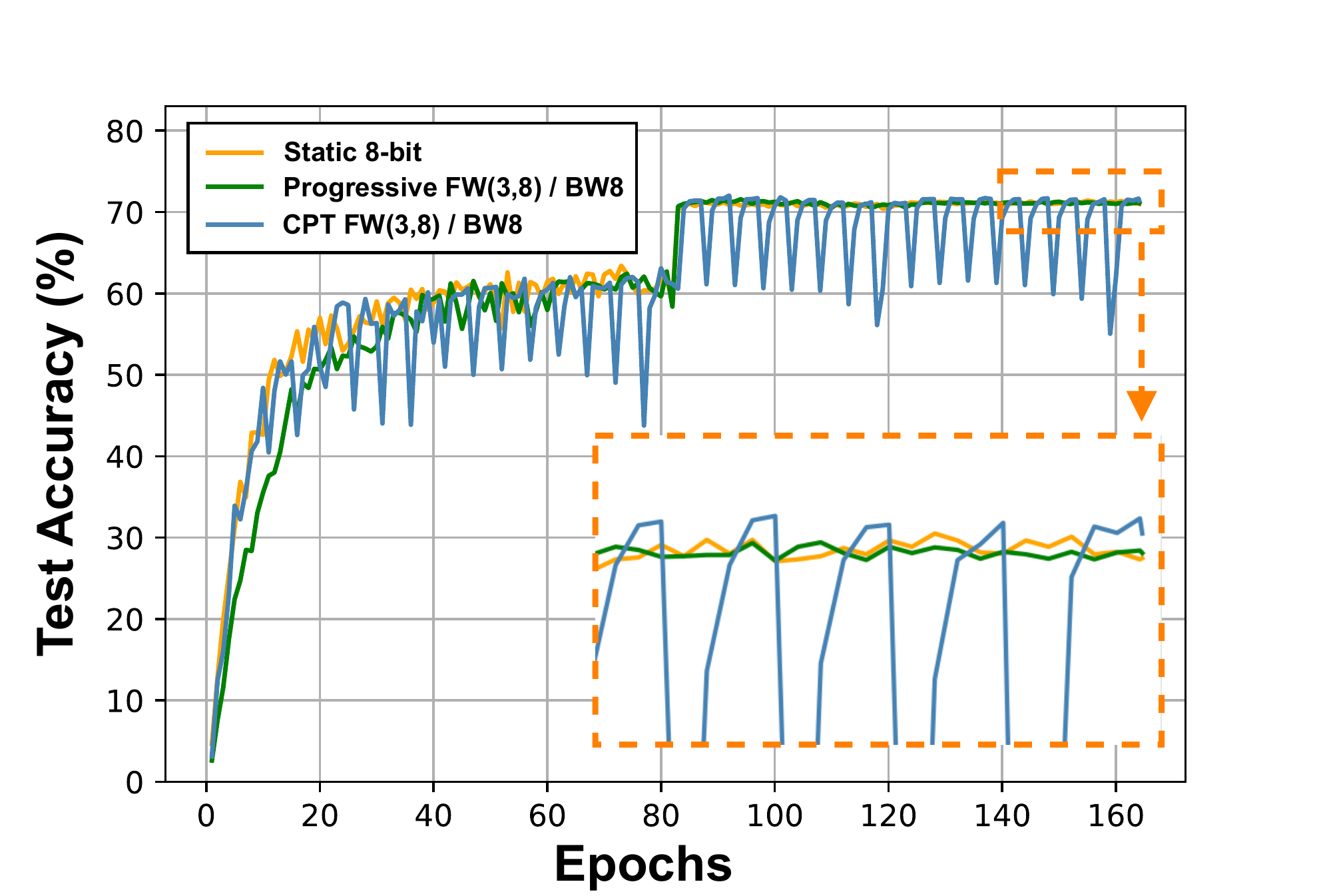}
    \vspace{-2.2em}
    \caption{Test accuracy evolution of ResNet-74 on CIFAR-100 under different schedules.}
    \label{fig:acc_progressive}
    \vspace{-1em}
% \end{figure*}
\end{wrapfigure}

\textbf{Hypothesis 2: Dynamic precision helps DNN generalization.} Recent findings in DNN training have motivated us to better utilize DNN precision to achieve a win-win in both DNN accuracy and efficiency. Specifically, it has been discussed that (1) DNNs learn to fit different patterns at different training stages, e.g., \citep{pmlr-v97-rahaman19a, xu2019frequency} reveal that DNN training first learns lower-frequency components and then high-frequency features, with the former being more robust to perturbations and noises; and (2) dynamic learning rate schedules help to improve the optimization in DNN training, e.g., ~\citep{li2019towards} points out that a large initial learning rate helps the model to memorize easier-to-fit and more generalizable patterns while ~\citep{smith2017cyclical, loshchilov2016sgdr} show that cyclical learning rate schedules improve DNNs' classification accuracy. These works inspire us to hypothesize that dynamic precision might help DNNs to reach a better optimum in the optimization landscape, especially considering the similar effect between the learning rate and precision validated in our Hypothesis 1.

\textbf{Validating Hypothesis 2.} Our Hypothesis 2 has been consistently confirmed by various empirical observations. For example, a recent work~\citep{fu2020fractrain} proposes to progressively increase the precision during the training process, and we follow their settings to validate our hypothesis.

\underline{Settings:} We train a ResNet-74 on CIFAR-100 using the same training setting as~\citep{wang2018skipnet} except that we quantize the weights, activations, and gradients during training; for the \textbf{progressive} precision case we uniformly increase the precision of weights and activations from 3-bit to 8-bit in the first 80 epochs and adopt static 8-bit gradients, while the static precision baseline uses 8-bit for all the weights/activations/gradients.

\begin{figure*}[!t]
    \centering
    %  \vspace{-1em}
    \includegraphics[width=\textwidth]{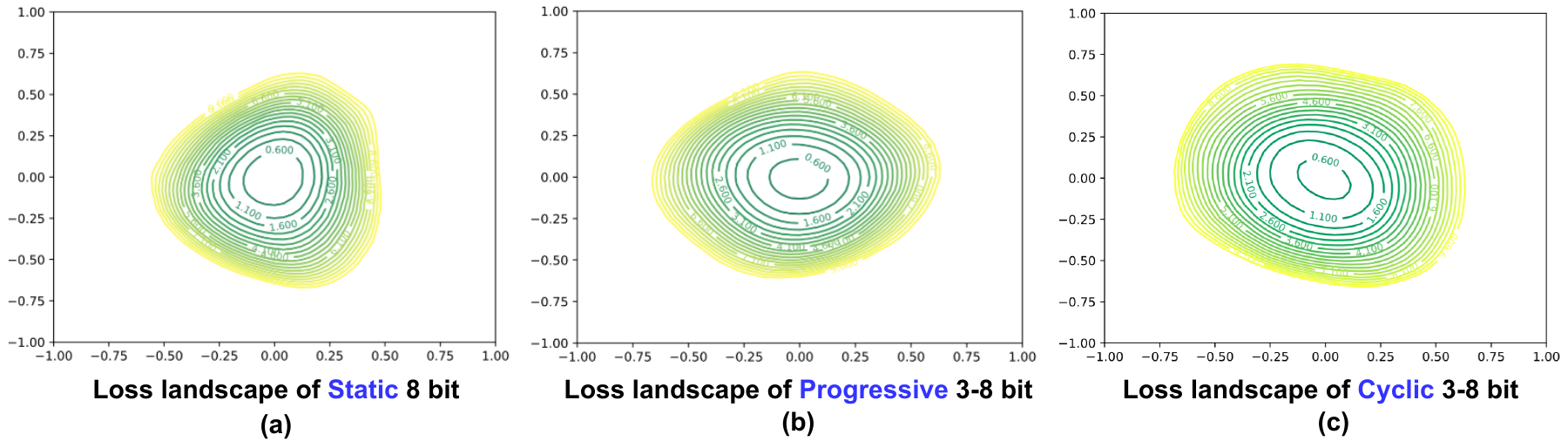}
     \vspace{-2em}
    \caption{Loss landscape visualization after convergence of ResNet-74 on CIFAR-100 trained with different precision schedules, where wider contours with larger intervals indicate a better local minima and a lower generalization error as analyzed in~\citep{li2018visualizing}.}
    \label{fig:motivation_loss_landscape}
    \vspace{-2em}
\end{figure*}

\underline{Results:} Fig.~\ref{fig:acc_progressive} shows that training with progressive precision schedule achieves a slightly higher accuracy (+0.3\%) than its static counterpart, while the former can reduce training costs. Furthermore, we visualize the loss landscape (following the method in~\citep{li2018visualizing}) in Fig.~\ref{fig:motivation_loss_landscape}(b): interestingly the progressive precision schedule helps to converge to a better local minima with wider contours, indicating a lower generalization error~\citep{li2018visualizing} over the static 8-bit baseline in Fig.~\ref{fig:motivation_loss_landscape}(a).
% smooth the loss surface and thus to learn more generalizable patterns with better convergence.

The progressive precision schedule in~\citep{fu2020fractrain} relies on manual hyper-parameter tuning. As such, a natural following question would be: what kind of dynamic schedules would be effective while being simple to implement for different tasks/models? In this work, we show that a simple cyclic schedule consistently benefits the training convergence while boosting the training efficiency.

\begin{wrapfigure}{r}{0.5\textwidth}
% \begin{figure*}[!t]
    \centering
    % \vspace{-2em}
    % \includegraphics[width=\textwidth]{figure/cyclic_overview.pdf}
     \vspace{-2em}
    \includegraphics[width=0.5\textwidth]{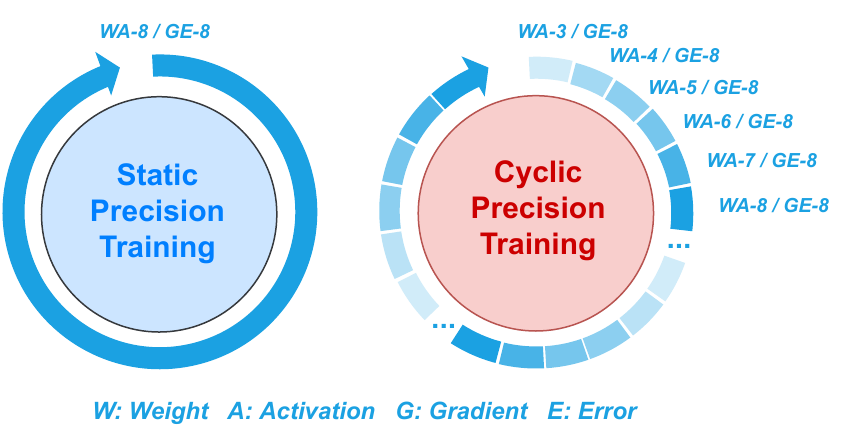}
    \vspace{-2.2em}
    \caption{Static vs. Cyclic Precision Training (CPT), where CPT cyclically schedules the precision of weights and activations during training.}
    \label{fig:cyclic_overview}
    \vspace{-1em}
% \end{figure*}
\end{wrapfigure}

\vspace{-0.5em}
\subsection{CPT: The Key Concept}
\label{sec:cyclic_precsion}
\vspace{-0.5em}
The key concept of CPT draws inspiration from \citep{li2019towards} which demonstrates that a large initial learning rate helps the model to learn more generalizable patterns. We thus hypothesize that a lower precision that leads to a short-term poor accuracy might actually help the DNN exploration during training thanks to its associated larger quantization noise, while it is well known that a higher precision enables the learning of higher-complexity, fine-grained patterns that is critical to better convergence. Together, this combination could improve the achieved accuracy as it might better balance coarse-grained exploration and fine-grained optimization during DNN training, which leads to the idea of CPT. Specifically, as shown in Fig.~\ref{fig:cyclic_overview}, CPT varies the precision cyclically between two bounds instead of fixing the precision during training, letting the models explore the optimization landscape with different granularities.

While CPT can be implemented using different cyclic scheduling methods, here we present as an example an implementation of CPT in a cosine manner:

\setlength{\abovedisplayskip}{0pt}
\setlength{\belowdisplayskip}{0pt}
\vspace{-0.5em}
\begin{align}
\vspace{-0.5em}
    \begin{split}
    B_t^n = \lceil B_{min}^{n} + \frac{1}{2}(B_{max}^{n} - B_{min}^{n})(1 - cos(\frac{t \, \% \, T_n}{T_n}\pi))\rfloor
    \label{eq:cyclic_precision}
    \end{split} 
\end{align}

where $B_{min}^{n}$ and $B_{max}^{n}$ are the lower and upper precision bound, respectively, in the $n$-th cycle of precision schedule, $\lceil \cdot \rfloor$ and $\%$ denote the rounding operation and the remainder operation, respectively, and $B_t^n$ is the precision at the $t$-th global epoch which falls into the $n$-th cycle with a cycle length of $T_n$. Note that the cycle length $T_n$ is equal to the total number of training epochs divided by the total number of cycles denoted as $N$, where $N$ is a hyper-parameter of CPT. For example, if $N=2$, then a DNN training with CPT will experience two cycles of cyclic precision schedule during training.
% , i.e., $\{T_n|n\in{0,1}\}=N/2$.
As shown in Sec.~\ref{sec:ablation}, we find that the benefits of CPT are maintained when adopting different total number of cyclic precision schedule cycles during training, i.e., CPT is not sensitive to $N$. 
% We leave the optimal cyclic scheduling method as a future work. 
\textcolor{black}{A visualization example for the precision schedule can be found in Appendix~\ref{appendix:visual}. Additionally, we find that CPT is generally effective when using different dynamic precision schedule patterns (i.e., not necessarily the cosine schedule in  Eq.~(\ref{eq:cyclic_precision}). We implement CPT following Eq.~(\ref{eq:cyclic_precision}) in this work and discuss the potential variants in Sec.~\ref{sec:ablation}.}

We visualize the training curve of CPT on ResNet-74 with CIFAR-100 in Fig.~\ref{fig:acc_progressive} and find that it achieves a 0.91\% higher accuracy paired with a 36.7\% reduction in the required training BitOPs (bit operations), as compared to its \textbf{static} fixed precision counterpart. In addition,
Fig.~\ref{fig:motivation_loss_landscape} (c) visualizes the corresponding loss landscape, showing the effectiveness of CPT, i.e., such a simple and automated precision schedule leads to a better convergence with lower sharpness.

\begin{wrapfigure}{r}{0.5\textwidth}
% \begin{figure*}[!t]
    \centering
    % \vspace{-2em}
    % \includegraphics[width=\textwidth]{figure/cyclic_overview.pdf}
     \vspace{-3em}
    \includegraphics[width=0.5\textwidth]{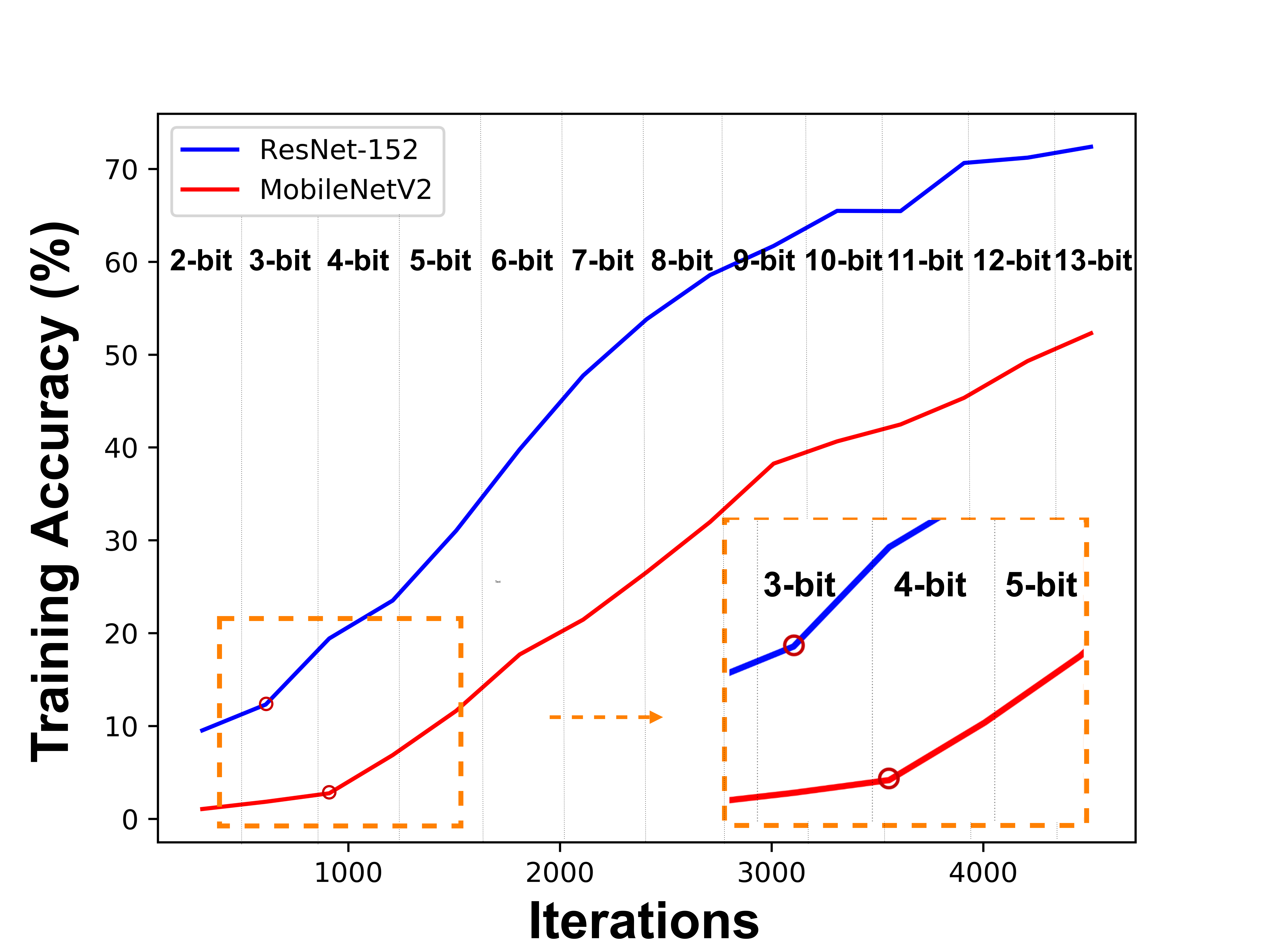}
    \vspace{-1.7em}
    \caption{Illustrating the precision range test for ResNet-152 and MobileNetV2 on CIFAR-100, where the switching point which exceeds the preset threshold is denoted by red circles.}
    \label{fig:range_test}
    \vspace{-1em}
% \end{figure*}
\end{wrapfigure}

% \vspace{-1.5em}
\subsection{CPT: Precision Range Test}
\label{sec:range_test}
\vspace{-0.5em}
The concept of CPT is simple enough to be plugged into any model or task to boost the training efficiency. One remaining question is how to determine the precision bounds, i.e., $B_{min}^{i}$ and $B_{max}^{i}$ in Eq.~(\ref{eq:cyclic_precision}), which we find can be automatically decided in the first cycle (i.e., $T_i=T_0$) of the precision schedule using a simple PRT at a negligible computational cost. Specifically, PRT starts from the lowest possible precision, e.g., 2-bit, and gradually increases the precision while monitoring the difference in the training accuracy magnitude averaged over several consecutive iterations; once this training accuracy difference is larger than a preset threshold, indicating that the training can at least partially converge, PRT would claim that the lower bound is identified. While the upper bound can be similarly determined, there exists an alternative which suggests simply adopting the precision of CPT's static precision counterpart. The remaining cycles use the same precision bounds.
% For example, if SOTA DNN quantization adopts 8-bit, one can simply plug in our CPT using 8-bit as the upper bound to further reduce  

Fig.~\ref{fig:range_test} visualizes the PRT for ResNet-152/MobileNetV2 trained on CIFAR-100. We can see that the lower precision bound identified when the model experiences a notable training accuracy improvement for ResNet-152 is 3-bit while that for MobileNetV2 is 4-bit, aligning with the common observation that ResNet-152 is more robust to quantization than the more compact model MobileNetV2.

% To fully automate the CPT training framework which is ready to plug into any model and task, the hyper-parameters in Eq.~\ref{eq:cyclic_precision} need to be automatically determined. Since there's a general applicable number of stages for all the tasks (Sec.~\ref{sec:ablation}), the main challenge is to determine lower bound and upper bound of precision range. Therefore, we propose a simple but effective technique name Precision Range Test (PRT) to adaptively adjust the precision bound with little overhead.

% PRT is conducted only during the first period that we start from the lowest possible precision, e.g., 2-bit, and gradually increase it as in Eq.~\ref{eq:cyclic_precision}. We keep monitoring the magnitude of loss difference averaging over several continuous iterations. When it's larger than a threshold, indicating under current precision the loss can be normally reduce, then the lower bound of precision range can be identified. Although the upper bound can be derived in a similar way, we calculate its value based on the target computational cost with known lower precision bound and the periodical schedule, considering the efficient training goal. After the first period, all the following periods will adopt the same derived precision range.

% Fig.~\ref{fig:XXX} visualizes the PRT of ResNet-74/MobileNet-V2 on CIFAR-100. We can see that the the lower bound with a notable loss reduction for ResNet-74 is 3-bit while that for MobileNetV2 is 4-bit, which aligns with the common observation that ResNet-74 is more robust to quantization than MobileNetV2.

%% file: Sections/4-Experiment.tex
\vspace{-0.3em}
\section{Experiment Results}
\label{sec:exp}
\vspace{-0.5em}
In this section, we will first describe the experiment setup in Sec.~\ref{sec:exp_setup}, benchmarking results over SOTA training methods across various tasks in Sec.~\ref{sec:exp_sota}, and then comprehensive ablation studies of CPT in Sec.~\ref{sec:exp_ablation}.
% In this section, we evaluate our CPT framework on \NOTE{eleven} models and \NOTE{six} tasks across classification, \NOTE{object detection} and language modeling. Sec.~\ref{sec:exp_setup} introduces the experiment setup. We then benchmark with SOTA methods across various tasks in Sec.~\ref{sec:exp_sota} and do a comprehensive ablation study of CPT in Sec.~\ref{sec:exp_ablation}.

\vspace{-0.5em}
\subsection{Experiment setup}
\label{sec:exp_setup}
\vspace{-1em}
\textbf{Models, datasets and baselines.} We consider \underline{\textcolor{black}{eleven} models} (including eight ResNet based models~\citep{he2016deep}, MobileNetV2~\citep{sandler2018mobilenetv2}, 
% Yolo~\citep{redmon2016you}, SSD~\citep{liu2016ssd}, 
Transformer~\citep{vaswani2017attention}, and LSTM~\citep{hochreiter1997long}) and \underline{\textcolor{black}{five} tasks} (including CIFAR-10/100~\citep{krizhevsky2009learning}, ImageNet~\citep{imagenet_cvpr09}, 
% COCO~\citep{lin2014microsoft},
WikiText-103~\citep{merity2016pointer}, and Penn Treebank (PTB)~\citep{marcus1993building}). Specifically, we follow~\citep{wang2019e2} for implementing MobileNetV2 on CIFAR-10/100. \underline{Baselines:} We first benchmark CPT over three SOTA static low-precision training techniques: SBM~\citep{banner2018scalable}, DoReFa~\citep{zhou2016dorefa}, and WAGEUBN~\citep{yang2020training}, each of which adopts a different quantizer. Since SBM is the most competitive baseline among the three based on both their reported and our experiment results, we apply CPT on top of SBM, and all the static precision baselines adopt the SBM quantizer unless specifically stated. Another baseline is the cyclic learning rate (CLR)~\citep{loshchilov2016sgdr} on top of static precision training, and we follow the best setting in~\citep{loshchilov2016sgdr}.
% \textbf{Models, datasets and baselines.} \underline{\NOTE{Eleven} models:} ResNet-18/34/38/74/110/152/164~\citep{he2016deep}, MobileNetV2~\citep{sandler2018mobilenetv2}, Yolo~\citep{redmon2016you}, SSD~\citep{liu2016ssd}, Transformer~\citep{vaswani2017attention}, LSTM~\citep{hochreiter1997long}. \underline{\NOTE{Six} tasks:} CIFAR-10/100~\citep{krizhevsky2009learning}, ImageNet~\citep{imagenet_cvpr09}, COCO~\citep{lin2014microsoft}, WikiText-103~\citep{merity2016pointer} and PTB~\citep{marcus1993building}. \underline{Baselines:} We benchmark with three SOTA quantizers for training, i.e., SBM~\citep{banner2018scalable}, DoReFa~\citep{zhou2016dorefa} and WAGEUBN~\citep{yang2020training}. Since SBM has the best performance among them, we build on top of their quantizer and all the static precision baselines use SBM unless specially annotated. Another baseline is cyclic learning rate (CLR) on top of static precision training, and we follow~\citep{smith2017cyclical} for the CLR settings.

\textbf{Training settings.} We follow the standard training setting in all the experiments. In particular, for classification tasks, we follow SOTA settings in~\citep{wang2018skipnet} for CIFAR-10/100 and~\citep{he2016deep} for ImageNet experiments, respectively; 
% \NOTE{for object detection experiments on COCO, we follow the settings in~\citep{XXX}}; 
and for language modeling tasks, we follow~\citep{baevski2018transformer} for Transformer on WikiText-103 and~\citep{merity2017regularizing} for LSTM on PTB. 
% \NOTE{More detailed training settings are provided in appendix.}

% \textbf{Training settings.} We follow the standard training setting in all the experiments. In particular, for classification tasks, we follow SOTA settings in~\citep{wang2018skipnet} for CIFAR-10/100 and~\citep{he2016deep} for ImageNet experiments. \NOTE{For object detection experiments on COCO, we follow~\citep{XXX}.} For language modeling tasks, we follow~\citep{XX} \NOTE{(Xin Yang add here)} for Transformer on WikiText-103 and~\citep{XX} for LSTM on PTB. \NOTE{All the detailed settings are clarified in appendix.}

\textbf{Precision settings.} The lower precision bounds in all the experiments are set using the PRT in Sec.\ref{sec:range_test} and the upper bound is the same as the precision of the corresponding static precision baselines. We only apply CPT to the weights and activations (together annotated as FW) and use static precision for the errors and gradients (together annotated as BW), the latter of which is to ensure the stability of the gradients~\citep{wang2018training} (more discussion in Appendix~\ref{appendix:gradient}). In particular, CPT from 3-bit to 8-bit with 8-bit gradient is annotated as FW(3,8)/BW8. 
% as discussed in Sec.~\ref{sec:ablation}. 
The total number of periodic precision  cycles, i.e., $N$ in Sec.\ref{sec:range_test}, for all the experiments is fixed to be 32 (see the ablation studies in Sec.~\ref{sec:exp_ablation}).

% \textbf{Precision settings.} The lower precision bound in all the experiments in are determined by PRT and the upper bound is determined by the user demand of target efficiency. We only apply CPT on weight and activation (annotated as FW) and use static precision for error and gradient (annotated as BW) due to the instability of gradients~\citep{wang2018training} which is also discussed in Sec.~\ref{sec:ablation}. The number of stages for all the experiments is fixed to be 32.

\textbf{Hardware settings and metrics.} To validate the real hardware efficiency of the proposed CPT, we adopt standard FPGA implementation flows. Specifically, we employ the Vivado HLx design flow to implement FPGA-based accelerators on a Xilinx development board called ZC706~\citep{zc706}.  To better evaluate the training cost, we consider both calculated  GBitOPs (Giga bit operations) and real-measured latency on the ZC706 FPGA board.

% \textbf{Efficiency metrics.} To better evaluate the training cost, we consider both calculated  GBitOPs (Giga bit operations) and real-measured latency on the Xilinx FPGA board ZC706~\citep{zc706}.
% We consider both theoretical complexity and real hardware cost for low precision training. We consider GBitOPs (Giga bit operations) for the former one and latency measure on ZC706 board for the latter one. 
%\textcolor{red}{Luke: citation for ZC706?}

\vspace{-1em}
\subsection{Benchmark with SOTA static precision training methods}
\label{sec:exp_sota}
\vspace{-0.5em}

\begin{figure}[!b]
    \centering
    % \vspace{-2em}
    \includegraphics[width=\textwidth]{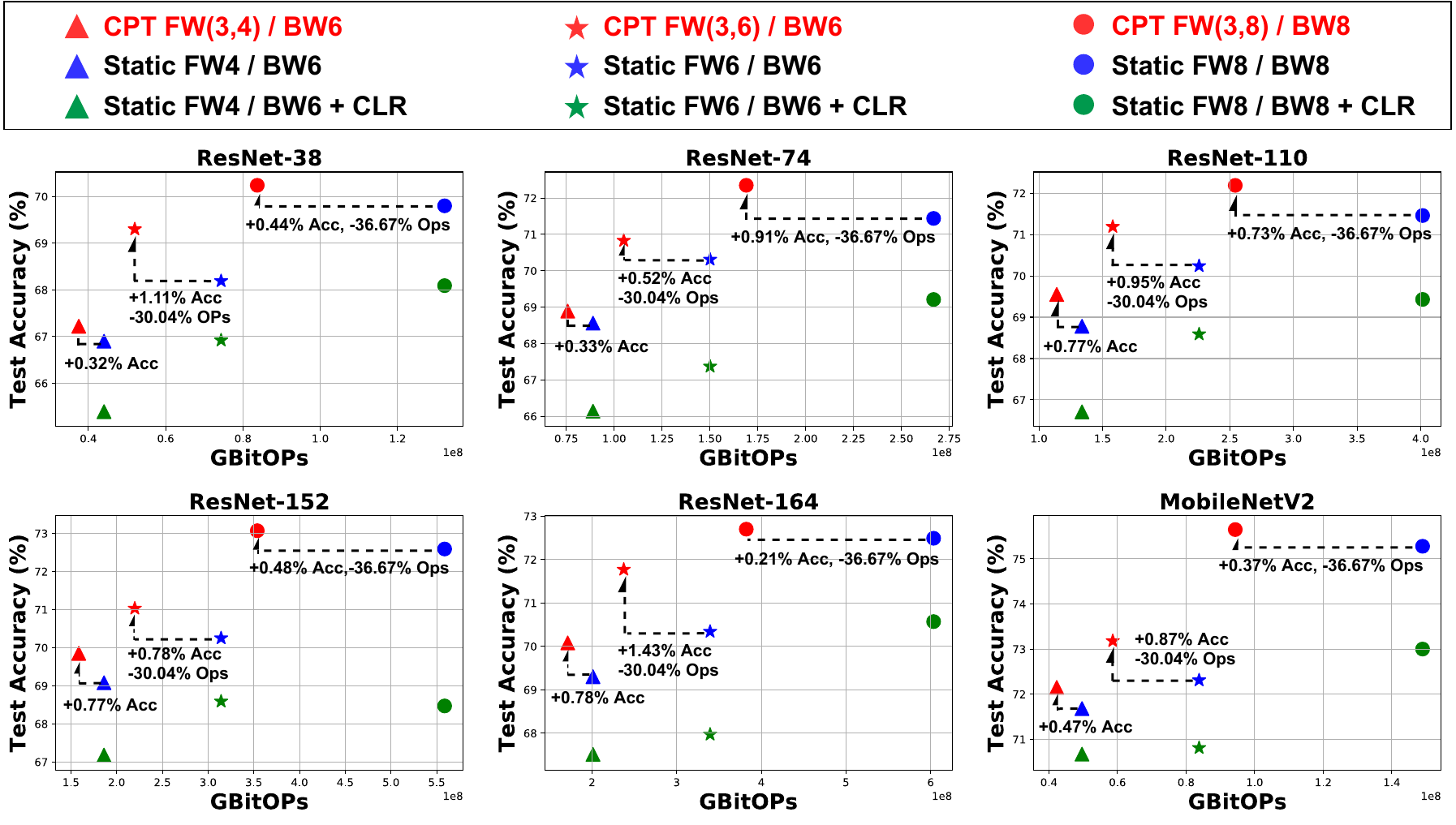}
    \vspace{-2em}
    \caption{Test accuracy vs. the required GBitOPs when training ResNet-38/74/110/152/164 and MobileNetV2 on CIFAR-100 using static precision, static precision plus CLR, and CPT methods.}
    \label{fig:benchmark_clr}
    %\vspace{-1em}
\end{figure}

\textbf{Benchmark on CIFAR-10/100.} 
\underline{Benchmark over SOTA quantizers:} We benchmark CPT with three SOTA static low-precision training methods, as summarized in Tab.~\ref{tab:sota_quantizer}, when training ResNet-74/164 and MobileNetV2 on CIFAR-10/100, covering both deep and compact DNNs which are representative difficult cases of low-precision DNN training. Note that \textbf{the accuracy improvement is the difference between CPT and the strongest baseline under the same setting}. 
% \NOTE{More results evaluated against SOTA quantizers on other DNN models are provided in appendix.} 
Tab.~\ref{tab:sota_quantizer} shows that (1) our CPT consistently achieves a win-win with both a higher accuracy (+0.19\% $\sim$ +1.25\%) and a lower training cost (-21.0\% $\sim$ -37.1\% computational cost and -14.7\% $\sim$ -21.4\% latency) under all the cases on CIFAR-10/100, even for the compact model MobileNetV2, and (2) CPT notably outperforms the baselines in terms of accuracy \textbf{under extremely low precision}, which is one of the most useful scenarios for low-precision DNN training. In particular, CPT with periodic precision between 4-bit and 6-bit boosts the accuracy by +1.25\% and +1.07\% on ResNet-74/164, respectively, as compared to the static 6-bit training on CIFAR-10, verifying that CPT leads to a better convergence.

\begin{table*}[!t]
\vspace{-1em}
  \centering
  \caption{The test accuracy, computational cost, and latency of CPT, DoReFa~\citep{zhou2016dorefa}, WAGEUBN~\citep{yang2020training}, and SBM~\citep{banner2018scalable} for training the ResNet-74/164 and MobileNetV2 models on CIFAR-10/100.}
  \vspace{-0.8em}
  \resizebox{\textwidth}{!}{ 
\begin{tabular}{ccccccc}
\toprule
\textbf{Model}  & \textbf{Method} & \textbf{Precision (FW/BW)} & \textbf{CIFAR-10 Acc (\%)} & \textbf{CIFAR-100 Acc (\%)} & \textbf{GBitOPs} & \textbf{Latency (hour)} \\

\midrule
\multirow{5}[4]{*}{ResNet-74} & DoReFa & 8 / 8  &  91.16     &   69.31    &   2.67e8 & 44.6  \\
       & WAGEUBN & 8 / 8  &   91.35     &   69.61     &   2.67e8     &  44.6  \\
       & SBM    & 8 / 8  &   92.57     &   71.44     &    2.67e8    &  44.6  \\
       & \textbf{Proposed CPT}    & \textbf{3 - 8 / 8} &    \textbf{93.23}    &   \textbf{72.35}     &  \textbf{1.68e8}   & \textbf{35.04}  \\
\cmidrule{2-7}       & \textbf{Improv.} &        &  \textbf{+0.66}      &   \textbf{+0.91}     &  \textbf{-37.1\%}      & \textbf{-21.4\%}  \\
\midrule
\multirow{5}[4]{*}{ResNet-74} & DoReFa & 6 / 6  &  90.94      &   69.01     &  1.50e8   & 33.2  \\
       & WAGEUBN & 6 / 6  &   91.01     &   69.37    &   1.50e8     &  33.2 \\
       & SBM    & 6 / 6  &    91.15    &   70.31     &  1.50e8      &  33.2  \\
       & \textbf{Proposed CPT}    & \textbf{3 - 6 / 6} &   \textbf{92.4}     &    \textbf{70.83}    &  \textbf{1.05e8}    & \textbf{27.5} \\
\cmidrule{2-7}       & \textbf{Improv.} &        &  \textbf{+1.25}      &   \textbf{+0.52}     &  \textbf{-30.0\%}      &  \textbf{-17.2}\%  \\
\midrule

\multirow{5}[4]{*}{ResNet-164} & DoReFa & 8 / 8  &   91.40     &   70.90     &   6.04e8   &  101.9 \\
       & WAGEUBN & 8 / 8  &   92.5     &    71.86    &   6.04e8     & 101.9 \\
       & SBM    & 8 / 8  &   93.63     &   72.53     &   6.04e8     & 101.9 \\
       & \textbf{Proposed CPT}    & \textbf{3 - 8 / 8} &   \textbf{93.83}     &    \textbf{72.9}    &   \textbf{3.8e8}    & \textbf{80.5} \\
\cmidrule{2-7}       & \textbf{Improv.} &        &    \textbf{+0.20}    &  \textbf{+0.37}      &  \textbf{-37.1\%}      &  \textbf{-21.0\%}  \\
\midrule
\multirow{5}[4]{*}{ResNet-164} & DoReFa & 6 / 6  &  91.13      &  70.53      &  3.40e8      &  76.7  \\
       & WAGEUBN & 6 / 6  &   92.44     &    71.50    &  3.40e8      & 76.7   \\
       & SBM    & 6 / 6  &    91.95    &     70.34   &  3.40e8  &  76.7 \\
       & \textbf{Proposed CPT}    & \textbf{3 - 6 / 6} &   \textbf{93.02}     &    \textbf{71.79}    &   \textbf{2.37e8}   &  \textbf{63.5} \\
\cmidrule{2-7}       & \textbf{Improv.} &   & \textbf{+1.07}      &  \textbf{+0.29}  &  \textbf{-30.3}\%  & \textbf{-17.2\%}   \\
\midrule
\multirow{5}[4]{*}{MobileNetV2} & DoReFa & 8 / 8  &   91.03     &  70.17      &   1.49e8     &  26.2 \\
       & WAGEUBN & 8 / 8  &    92.32    &    71.45    &  1.49e8      & 26.2 \\
       & SBM    & 8 / 8  &    93.57    &     75.28   &  1.49e8      &  26.2 \\
       & \textbf{Proposed CPT}    & \textbf{4 - 8 / 8} &    \textbf{93.76}    &   \textbf{75.65}     &   \textbf{1.04e8}   &  \textbf{21.6} \\
\cmidrule{2-7}       & \textbf{Improv.} &        &  \textbf{+0.19}      &  \textbf{+0.37}      &  \textbf{-30.2\%}      & \textbf{-17.6\%}   \\
\midrule
\multirow{5}[4]{*}{MobileNetV2} & DoReFa & 6 / 6  &    90.25    &   68.4     &   8.39e7     &  18.4 \\
       & WAGEUBN & 6 / 6  &    91.00    &   71.05     &  8.39e7      &  18.4  \\
       & SBM    & 6 / 6  &    91.56    &     72.31   &   8.39e7     & 18.4 \\
       & \textbf{Proposed CPT}    & \textbf{4 - 6 / 6} &    \textbf{91.81}    &    \textbf{73.18}    &  \textbf{6.63e7}      &  \textbf{15.7} \\
\cmidrule{2-7}       & \textbf{Improv.} &        &  \textbf{+0.25}     & \textbf{+0.87}       &  \textbf{-21.0\%}      &  \textbf{-14.7\%} \\
\bottomrule
\end{tabular}%
    }

  \label{tab:sota_quantizer}%
  \vspace{-1em}
\end{table*}%

% \textbf{Benchmark on CIFAR-10/100.} 
% \underline{Benchmark with different quantizers:} We benchmark with three SOTA static precision training methods in Tab.~\ref{tab:sota_quantizer} of ResNet-74/164 and MobileNetV2 on CIFAR-10/100, covering both deep and compact models which are hard cases for low precision training. \NOTE{More results on different quantizers on other networks are in appendix.} From Tab.~\ref{tab:sota_quantizer} we can observe that (1) our CPT consistently achieves a win-win with both higher accuracy (\NOTE{+XX\% $\sim$ +XX\%}) and efficiency (\NOTE{-XX\% $\sim$ -XX\%} computational cost and \NOTE{+XX\% $\sim$ +XX\%} latency) under all the cases on CIFAR-10/100, even for the compact model. (2) CPT outperforms the baselines in terms of accuracy especially under lower precision. In particular, CPT with periodic precision between 4-bit and 6-bit improves \NOTE{+XX\% $\sim$ +XX\%} compared with static 6-bit on CIFAR-100 dataset, which verifies CPT leads to better convergence.

% We further benchmark with the best-performed baseline SBM on more models and precision in Fig.~\ref{fig:benchmark_clr} to find CPT still consistently outperforms the baselines with a better accuracy and efficiency trade-off.

\underline{Benchmark over CLR on top of SBM:} We further benchmark CPT with CLR~\citep{loshchilov2016sgdr} (inherit its besting setting on CIFAR-100), with both being applied on top of SBM as it achieves the best performance among SOTA quantizers as shown in Tab.~\ref{tab:sota_quantizer}, based on more DNN models and precision as shown in Fig.~\ref{fig:benchmark_clr}. We can see that (1) CPT still consistently outperforms all the baselines with a better accuracy and efficiency trade-off, and (2) CLR on top of SBM leads to a negative effect on the test accuracy, which we conjecture is caused by both the instability of gradients and the sensitivity of gradients to the learning rate under low-precision training, as discussed in~\citep{wang2018training}, showing that CPT is more applicable to low-precision training than CLR.

\begin{figure*}[!b]
    \centering
    \vspace{-1em}
    \includegraphics[width=\textwidth]{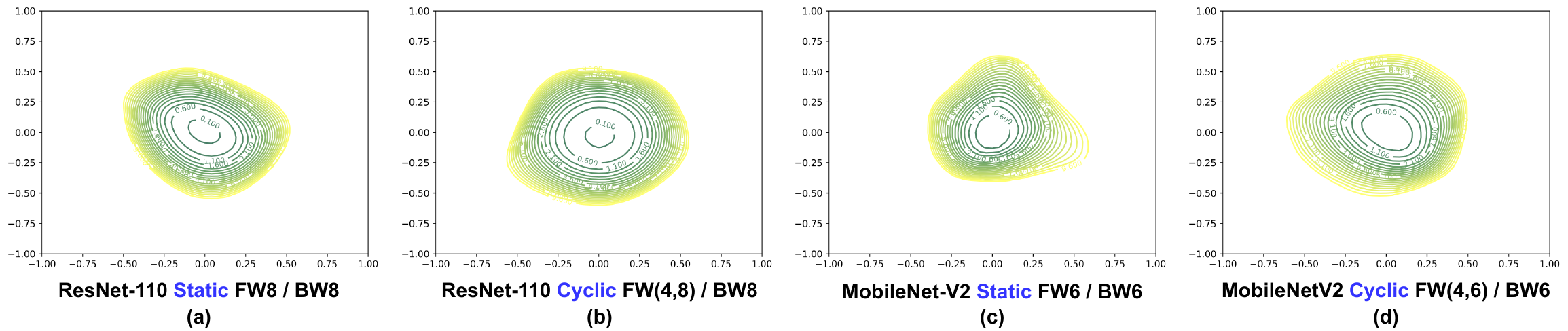}
    \vspace{-2em}
    \caption{Loss landscape visualization of ResNet-110 and MobileNetV2 trained on CIFAR-100.}
    \label{fig:landscape_exp}
    \vspace{-1em}
\end{figure*}

\underline{Loss landscape visualization:} To better understand the superior performance achieved by CPT, we visualize the loss landscape following the method in~\citep{li2018visualizing}, as shown in Fig.~\ref{fig:landscape_exp} covering both non-compact and compact models and two low-precision settings (i.e., 6-bit and 8-bit) which are bottlenecks for low-precision DNN training. We can observe that, again, both standard and compact DNNs trained with CPT experience wider contours with less sharpness, indicating that CPT helps DNN training optimization to converge to better local optima.

\begin{table*}[!th]
% \vspace{-0.8em}
  \centering
  \caption{The test accuracy and computational cost of ResNet-18/34/50 on ImageNet, trained with the proposed CPT and SBM~\citep{banner2018scalable}.}
  \vspace{-0.8em}
  \resizebox{0.9\textwidth}{!}{ 
\begin{tabular}{cccccccc}
\toprule
\textbf{Model}  & \textbf{Method} & \textbf{Precision (FW/BW)} & \textbf{Acc (\%)} & \textbf{GBitOPs} & \textbf{Precision} & \textbf{Acc (\%)} & \textbf{GBitOPs} \\
\midrule
\multirow{3}[4]{*}{ResNet-18} & SBM    & 8 / 8  &  69.60    &   2.86e9     & 6 / 6  &   69.30     &  1.61e9    \\
       & \textbf{Proposed CPT}    & 4 - 8 / 8 &    \textbf{69.64}    &   \textbf{1.99e9}     & 4 - 6 / 6 &    \textbf{69.33}   & \textbf{1.27e9}   \\
\cmidrule{2-8}       & \textbf{Improv.} &        &  \textbf{+0.04}      &  \textbf{-30.4}\%  &    & \textbf{+0.03}        & \textbf{-21.1}\%  \\
\midrule
\multirow{3}[4]{*}{ResNet-34} & SBM    & 8 / 8  &    \textbf{73.32}    &    5.77e9    & 6 / 6  &   \textbf{72.82}     & 3.24e9  \\
       & \textbf{Proposed CPT}    & 4 - 8 / 8 &   73.12     &   \textbf{4.03e7}     & 4 - 6 / 6 &    72.74    &  \textbf{2.57e9} \\
\cmidrule{2-8}       & \textbf{Improv.} &        &  -0.20      &  \textbf{-30.2}\%  &    &   -0.08     &      \textbf{-20.7}\%  \\
\midrule
\multirow{3}[4]{*}{ResNet-50} & SBM    & 8 / 8  &   76.29     &   6.47e9     & 6 / 6  &   75.72     & 3.63e9  \\
       & \textbf{Proposed CPT}    & 4 - 8 / 8   & \textbf{76.35}    &   \textbf{4.51e9}     & 4 - 6 / 6 &    \textbf{75.74}    & \textbf{2.87e9}  \\
\cmidrule{2-8}       & \textbf{Improv.} &        &   \textbf{+0.06}     &  \textbf{-30.3}\%      &        &   \textbf{+0.02}     &  \textbf{-20.9}\% \\
\bottomrule
\end{tabular}%
    }
  \label{tab:imagenet}%
  \vspace{-1em}
\end{table*}%

\textbf{Benchmark on ImageNet.}
To verify the scalability of CPT on more complex tasks and larger models, we benchmark CPT with the SOTA static precision training method SBM~\citep{banner2018scalable} on ResNet-18/34/50 with ImageNet, under which it is challenging for low-precision training such as 4-bit to work. As shown in Tab.~\ref{tab:imagenet}, we can observe that CPT still achieves a reduced computational cost (up to -30.4\%) with a comparable accuracy (-0.20\% $\sim$ +0.06\%). In particular, CPT works well on ResNet-50, leading to both a slightly higher accuracy and a better training efficiency, indicating the scalability of CPT with model complexity, and thus, its potential application in large scale training, in addition to on-device training scenarios.
% To verify the scalability of CPT on more complex tasks and larger models, we benchmark CPT with SOTA static precision training method SBM~\citep{banner2018scalable} on ResNet-18/34/50 and ImageNet, which is harder for low precision training such as 4-bit to work. As shown in Tab.~\ref{tab:imagenet}, we can observe that CPT still achieves comparable accuracy (\NOTE{+XX\% $\sim$ +XX\%}) and reduced computation (up to \NOTE{+XX\%}). In particular, CPT works well on ResNet-50 with both slight higher accuracy and better efficiency, indicating the scalability of CDT with model complexity and the potential application in large scale training in addition to on-site training.

\begin{wraptable}{r}{0.45\textwidth}
\vspace{-1em}
\caption{The test accuracy of ResNet-18/34 on ImageNet: CPT (8-32) vs. full precision.}
\label{tab:full_precision_imagenet}
\vspace{-0.5em}
\resizebox{0.45\textwidth}{!}{
\begin{tabular}{cccc}
\hline
\textbf{Network} & \textbf{Method} & \textbf{Precision} & \textbf{Acc (\%)} \\ \hline
\multirow{3}{*}{ResNet-18} & Full Precision & 32 & 69.76 \\
 & \textbf{Proposed CPT} & 8-32 & \textbf{70.67} \\ \cline{2-4} 
 & \textbf{Improv.} &  & \textbf{+0.91} \\ \hline
\multirow{3}{*}{ResNet-34} & Full Precision & 32 & 73.30 \\
 & \textbf{Proposed CPT} & 8-32 & \textbf{74.14} \\ \cline{2-4} 
 & \textbf{Improv.} &  & \textbf{+0.84} \\ \hline
\end{tabular}
}
\vspace{-1em}
\end{wraptable}

\underline{CPT for boosting accuracy:} 
An important perspective of CPT is its potential to improve training optimality in addition to efficiency. We illustrate CPT's advantage in improving the final accuracy through training ResNet-18/34 on ImageNet using CPT and static full precision. As shown in Tab.~\ref{tab:full_precision_imagenet}, CPT on ResNet-18/34 achieves a 0.91\%/0.84\% higher accuracy than their full precision counterparts on ImageNet, indicating that CPT can be adopted as a general technique to improve the final accuracy in addition to efficient training.

\textbf{Benchmark on WikiText-103 and PTB.} 
We also apply CPT on language modeling tasks (including WikiText-103 and PTB) (see Tab.~\ref{tab:language}) to show that CPT is also applicable to natural language processing models. Tab.~\ref{tab:language} shows that (1) CPT again consistently achieves a win-win in terms of accuracy (i.e., \textcolor{black}{perplexity} - the lower the better) and training efficiency, and (2) language modeling models/tasks are more sensitive to quantization, especially in LSTM models, as it always adapts to a larger lower precision bound, which is consistent with SOTA observations \citep{hou2019normalization}.

\begin{table*}[!th]
\vspace{-0.5em}
  \centering
  \caption{The test accuracy and computational cost of (1) Transformer on WikiText-103 and (2) 2-LSTM (two-layer LSTM) on PTB, trained with CPT and SBM~\citep{banner2018scalable}.}
  \vspace{-0.8em}
  \resizebox{0.9\textwidth}{!}{ 
\begin{tabular}{cccccccc}
\toprule
\textbf{Model / Dataset} & \textbf{Method} & \textbf{Precision (FW/BW)} & \textbf{Perplexity} & \textbf{GBitOPs} & \textbf{Precision} & \textbf{Perplexity} & \textbf{GBitOPs} \\
\midrule
\multirow{3}[4]{*}{\tabincell{c}{Transformer\\WikiText-103}} & SBM    & 8 / 8  &   31.77     &  1.44e6      & 6 / 8  &    32.41    &  9.87e5 \\
       & \textbf{Proposed CPT}    & \textbf{4 - 8 / 8} &   \textbf{30.22}     &    \textbf{1.0e6}    &\textbf{ 4 - 6 / 8} &  \textbf{ 31.0}    &  \textbf{7.66e5} \\
\cmidrule{2-8}       & \textbf{Improv.} &        &   \textbf{ -1.55}    & \textbf{  -30.2\%}     &        & \textbf{ -1.41 }     &  \textbf{-22.4\%}  \\
\midrule
\multirow{3}[4]{*}{\tabincell{c}{2-LSTM\\PTB}} & SBM    & 8 / 8  &    96.95 & 4.03e3        & 6 / 8  &     97.47   &   2.77e3 \\
       & \textbf{Proposed CPT }   & \textbf{5 - 8 / 8} &   \textbf{96.39 }    &  \textbf{3.09e3}      &\textbf{ 5 - 6 / 8} &   \textbf{ 97.0 }   &  \textbf{2.48e3} \\
\cmidrule{2-8}       & \textbf{Improv.} &        & \textbf{-0.56 }    & \textbf{ -23.2\%} &     & \textbf{-0.47 }    &  \textbf{-10.5\%}  \\
% \midrule
% \multirow{3}[4]{*}{\tabincell{c}{4-LSTM\\PTB}} & SBM    & 8 / 8  &        &        & 6 / 8  &        &  \\
%       & Proposed CPT    & 5 - 8 / 8 &        &        & 5 - 6 / 8 &        &  \\
% \cmidrule{2-8}       & Improv. &        &        &        &        &        &  \\
\bottomrule
\end{tabular}%
    }
  \label{tab:language}%
  \vspace{-0.5em}
\end{table*}%

% \textbf{Benchmark on COCO for Object Detection.} \NOTE{(On-going, cannot guarantee the final result.)}

\vspace{-0.5em}
\subsection{Ablation Studies of CPT}
\label{sec:exp_ablation}
\label{sec:ablation}
\vspace{-0.5em}

\begin{wrapfigure}{r}{0.4\textwidth}
% \begin{figure*}[!t]
    \centering
    \vspace{-2em}
    \includegraphics[width=0.35\textwidth]{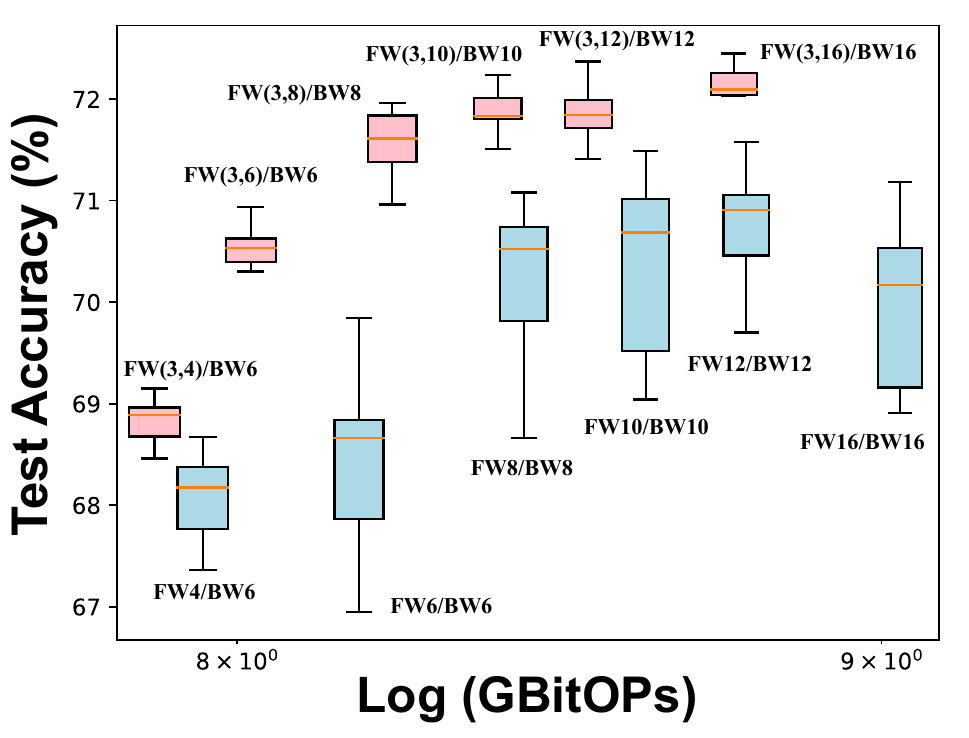}
    \vspace{-1em}
    \caption{Training ResNet-74 on CIFAR-100 with CPT and its static counterpart.}
    \label{fig:precision_range}
    \vspace{-2em}
% \end{figure*}
\end{wrapfigure}

\textbf{CPT with different precision ranges.} 
We also evaluate CPT under a wide range of upper precision bounds, which correspond to a different target efficiency, to see if CPT still works well. Fig.~\ref{fig:precision_range} plots a boxplot with each experiment being repeated ten times. We can see that
(1) regardless of the adopted precision ranges, CPT consistently achieves a win-win (a +0.74\% $\sim$ +2.03\% higher accuracy and a -18.3\% $\sim$ -48.0\% reduction in computational cost), especially under lower precision scenarios, and (2) \textbf{CPT even shrinks the accuracy variance}, which better aligns with the practical goal of efficient training.

\begin{figure*}[h]
    \centering
    % \vspace{-2em}
    \includegraphics[width=0.7\textwidth]{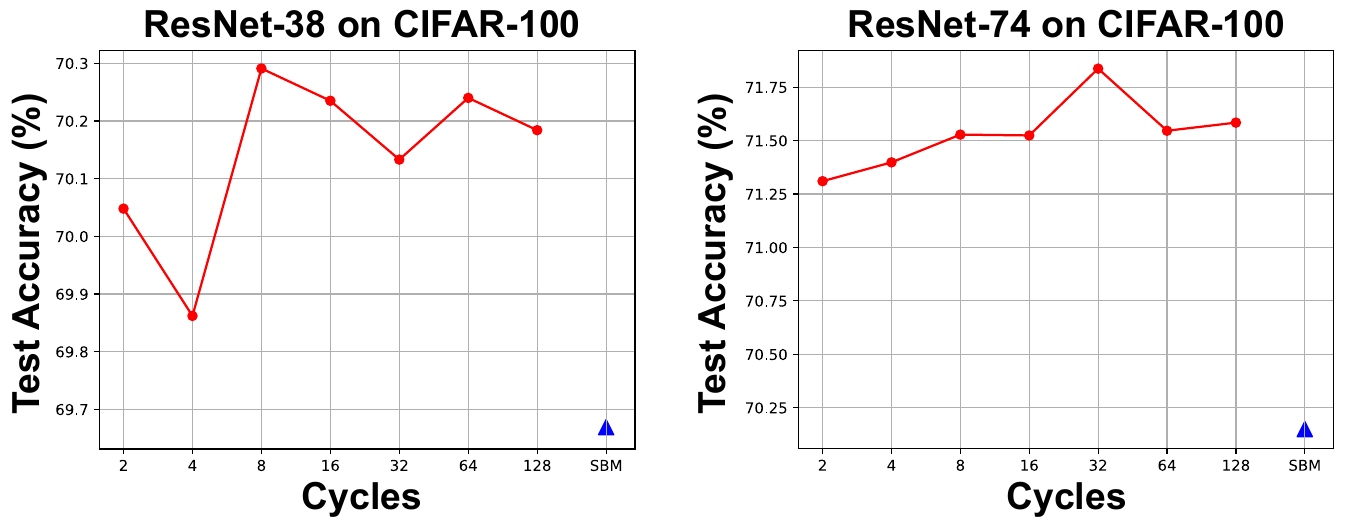}
   \vspace{-1em}
    \caption{The achieved test accuracy of CPT under different adopted numbers of precision schedule cycles, as compared to the static precision baseline SBM, when training ResNet-38/74 on CIFAR-100.}
    \label{fig:ablation_cycle}
    \vspace{-1.5em}
\end{figure*}

\textbf{CPT with different adopted numbers of precision schedule cycles.}
To evaluate CPT's sensitivity to the total number of adopted cyclic precision schedule cycles, we apply CPT on ResNet-38/74 and CIFAR-100 when using different numbers of schedule cycles. Fig.~\ref{fig:ablation_cycle} plots the mean of the achieved accuracy based on ten repeated experiments. We can see that (1) different choices for the total number of cycles lead to a comparable accuracy (within 0.5\% accuracy) on CIFAR-10/100;
% , so long as they are larger than a threshold number; 
and (2) CPT with different number of precision schedule cycles consistently outperforms the static baseline SBM in the achieved accuracy. Based on this experiment, we set $N=32$ for simplicity.
% \textcolor{red}{you don't mention random runs here, when does the mean and variance come from?}

\textbf{CPT with different cyclic precision schedule patterns.} We also evaluate CPT using other cyclic precision schedule patterns in addition to the cosine one, including a triangular schedule motivated by~\citep{smith2017cyclical} and a cosine annealing schedule as the learning rate schedule in~\citep{loshchilov2016sgdr}, with all adopting 32 cyclic cycles for fairness. Experiments in Tab.~\ref{tab:other_schedule} show that (1) CPT with different schedule patterns is consistently effective, and (2) CPT with the other two schedule patterns even surpasses the cosine one in some cases but underperforms on compact models. 
We leave how to determine the optimal cyclic pattern for a given model and task as a future work.

\begin{table*}[!th]
\vspace{-0.8em}
  \centering
  \caption{CPT with different precision schedules into cyclic precision training for ResNet-74/164 and MobileNetV2 on CIFAR-100. Cosine (CPT) is the current schedule adopted by CPT.}
  \vspace{-0.5em}
  \resizebox{0.9\textwidth}{!}{ 
\begin{tabular}{ ccccccc }
\toprule
	 & \multicolumn{2}{c}{\textbf{ResNet-74}}  & \multicolumn{2}{c}{\textbf{ResNet-164}} & \multicolumn{2}{c}{\textbf{MobileNetV2}}  \\ 
	 \midrule
	Schedule & FW(3,8) / BW8 & FW(3,6) / BW6 & FW(3,8) / BW8 & FW(3,6) / BW6 & FW(4,8) / BW8 & FW(4,6) / BW6 \\
    \midrule
% 	\cmidrule{2-7}
	Cosine (our CPT) & \textbf{72.35} & 70.83 & 72.7 & 71.77 & \textbf{75.65} & \textbf{73.18}  \\ 
% 	\cmidrule{2-7}
	Triangular & 71.61 & \textbf{71.22} & \textbf{72.94} & \textbf{72.37} & 74.92 & 71.37   \\ 
% 	\cmidrule{2-7}
	Cosine anneal & 71.37 & 70.80 & 72.4 & 72.03 & 74.59 & 73.12 \\ 
	\bottomrule
\end{tabular}
}
  \label{tab:other_schedule}%
  \vspace{-0.5em}
\end{table*}%

% \textbf{CPT on top of the DoReFa quantizer.}
% To evaluate CPT's sensitivity to the adopted quantizer, we further apply CPT on top of the DoReFa ~\citep{zhou2016dorefa} quantizer trained on ResNet-74 and CIFAR-100 as shown in Tab.~\ref{XXX}. We can observe that CPT again achieves both a better accuracy and a reduced training cost, verifying the general applicability of CPT regardless of the adopted quantizer.

% \textbf{CPT on gradients.}
% Here we apply CPT to the gradients (in addition to the weights and activations) to see if it can still work well. As shown in Tab.~\ref{XX}, CPT on gradients achieves an inferior accuracy compared with CPT applied merely to the weights and activations, which is consistent with the instability of gradients \textcolor{black}{under low precision training} discussed in~\citep{wang2018training}. 

\section{Discussions about future work}
\vspace{-0.5em}

\textbf{The theoretical perspective of CPT.}
There has been a growing interest in understanding and optimizing DNN training. For example, \citep{li2019towards} shows that training DNNs with a large initial learning rate helps the model to memorize more generalizable patterns faster and better. 
Recently,~\citep{zhu2020towards} showed that under a convexity assumption the convergence bound of reduced-precision DNN training is determined by a linear combination of the quantization noise and learning rate. These findings regarding DNN training seem to be consistent with the effectiveness of our CPT.
% In fact,~\citep{zhu2020towards} motivates us to rethink the role of the adopted precision in DNN training and propose CPT. 
% We will explore the theoretical perspective behind the success of CPT as our next step.

\textbf{The hardware support for CPT.} 
Recent progresses in mixed-precision DNN accelerators~\citep{lee20197, kim20201b} with dedicated modules for supporting dynamic precisions are promising to support our CPT. We leave the optimal accelerator design for CPT as our future works.

% There have been growing interests in understanding and optimizing DNN training. For example, \citep{pmlr-v97-rahaman19a,xu2019frequency} advocate that DNN training first learns low-complexity (lower-frequency) functional components and then high-frequency features, with the former being less sensitive to perturbations; \citep{achille2018critical} argues that important connections and the connectivity patterns between layers are first discovered at the early stage of DNN training, and then becomes relatively fixed in the latter training stage, which seems to indicate that critical connections can be learned independent of and also ahead of the final converged weights; and \citep{li2019towards} shows that training DNNs with a large initial learning rate helps the model to memorize easier-to-fit and more generalizable pattern faster and better. Those findings regarding DNN training seem to be consistent with the effectiveness of our proposed FracTrain. 

%% file: Sections/5-Conclusion.tex
\vspace{-0.8em}
\section{Conclusion}
\vspace{-0.8em}
We hypothesize that DNNs' precision has a similar effect as the learning rate when it comes to DNN training, i.e., low precision with large quantization noise helps DNN training exploration while high precision with more accurate updates aids model convergence, and thus advocate that dynamic precision schedules help DNN training optimization. We then propose the CPT framework which adopts a periodic precision schedule for low-precision DNN training in order to boost the achievable Pareto frontier of task accuracy and training efficiency. 
% Furthermore, we show that the periodic precision bounds can be automatically identified at the very early training stage using a simple precision range test, which has negligible computational overhead.
Extensive experiments and ablation studies verify that CPT can reduce computational cost during training while achieving a comparable or even better accuracy. Our future work will strive to identify more theoretical grounds for such dynamic low-precision training. 

\vspace{-0.8em}
\section*{Acknowledgement}
\vspace{-0.8em}
The work is supported by the National Science Foundation (NSF) through the Real-Time Machine Learning (RTML) program (Award number: 1937592).

%% file: Sections/6-Appendix.tex
\appendix
\section{Visualization of the precision schedule in CPT}
\label{appendix:visual}
\begin{figure*}[!t]
    \centering
    % \vspace{-2em}
    \includegraphics[width=0.8\textwidth]{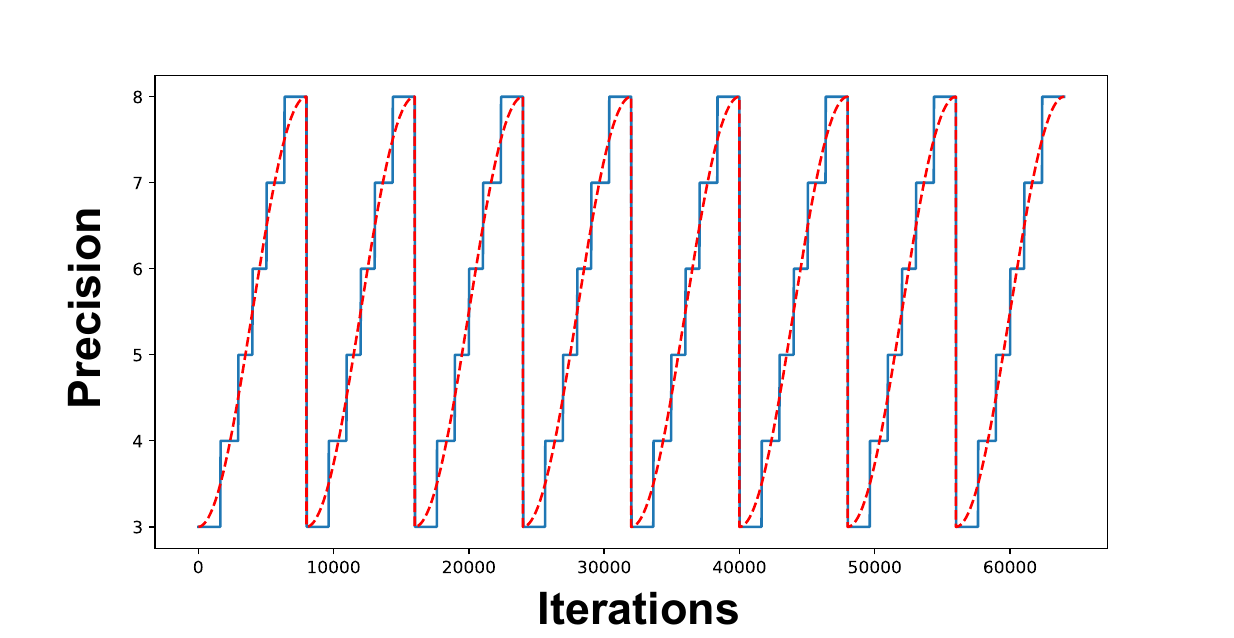}
    % \vspace{-1em}
    \caption{Visualization of the precision schedule FW(3,8) with eight cycles, where the red line is the cosine-manner schedule and the blue line is the adopted precision after rounding.}
    \label{fig:precision_schedule}
    % \vspace{-1em}
\end{figure*}

Fig.~\ref{fig:precision_schedule} visualizes the precision schedule FW(3,8) with eight cycles for training on CIFAR-10/100.

\begin{table}[h]
\centering
\caption{The test accuracy of ResNet-38/74 on CIFAR-10 trained with CPT enabled at different epochs. Note that the lowest training precision is applied before enabling CPT in all the experiments.}
\begin{tabular}{cccccccc}
\hline
Network & Starting Epoch & 0 & 60 & 80 & 100 & 120 & 160 (No CPT) \\ \hline
\multirow{2}{*}{ResNet-38} & FW(3,8) / BW8 & \textbf{70.24} & 69.60 & 68.37 & 68.04 & 67.65 & 62.3 \\
 & FW(3,6) / BW6 & \textbf{69.30} & 68.65 & 68.00 & 67.83 & 67.85 & 62.3 \\ \hline
\multicolumn{1}{l}{\multirow{2}{*}{ResNet-74}} & FW(3,8) / BW8 & \textbf{72.35} & 70.55 & 69.44 & 69.19 & 68.96 & 64.2 \\
\multicolumn{1}{l}{} & FW(3,6) / BW6 & \textbf{70.83} & 70.32 & 69.25 & 68.72 & 68.56 & 64.2 \\ \hline
\end{tabular}
\label{tab:starting_epoch}
\end{table}

\section{Ablation study: The starting epoch for enabling CPT}
We conduct another ablation study to explore whether CPT is always necessary during the whole training process. We start from training with the lowest precision in the precision range of CPT, and then enable CPT at different epochs to see the influence on the final accuracy.
Specifically, we train ResNet-38/74 on CIFAR-100 for 160 epochs, considering two CPT precision ranges with different starting epochs to enable CPT as shown in Tab.~\ref{tab:starting_epoch}. We consistently observe that (1) an early starting epoch of CPT leads to a better accuracy on all the considered precision ranges and networks, and (2) even enabling CPT at a later training stage still leads to a notable better accuracy than training without CPT.

% \begin{wraptable}{r}{0.4\textwidth}
\begin{wraptable}{r}{0.35\textwidth}
  \vspace{-1em}
  \centering
  \caption{Training ResNet-74 on CIFAR-100 with static precision and CPT on top of gradient only.}
  \vspace{-0.5em}
    \resizebox{0.3\textwidth}{!}{
\begin{tabular}{ccc}
\toprule
	FW & BW & Accuracy (\%)  \\
    \midrule
    6  & 6 & 70.31 \\
    
	6 & 8 & 70.51  \\ 
	
	\midrule
	\midrule

	6 & 6-8 &  69.45  \\ 

	6 & 6-10 &  70.32  \\ 
	
	6 & 6-12 &  69.59  \\ 
	
	6 & 6-16 &  70.24   \\
\bottomrule
\end{tabular}
    }
  \label{tab:gradient}%
  \vspace{-2em}
% \end{wraptable}%
\end{wraptable}%

\section{Ablation study: CPT on top of gradients}
\label{appendix:gradient}
We decide not to apply CPT on top of gradients since (1) the resulting instability of low precision gradients during training~\citep{wang2018training} can harm the final convergence, and (2) generally the required precision of gradients is higher than that of weights and activations, so that the benefit of applying CPT on top of gradients in terms of efficiency is limited.

To validate this, we show the results of applying CPT on top of gradients with fixed precision for weights and activations in Tab.~\ref{tab:gradient}. As expected, CPT on top of gradients can hardly benefit either accuracy or efficiency as compared with its static precision baseline. Therefore, we decide to adopt fixed precision for gradients in all other experiments.